\documentclass[twocolumn,10pt]{IEEEtran}
\usepackage{amsmath,epsfig,multirow,booktabs,amssymb,cite,mathtools}
\newcommand{\argmin}{\operatornamewithlimits{argmin}}

\newcommand{\defeq}{\vcentcolon=}

\begin{document}
\title{Multiple Kernel Sparse Representations for Supervised and Unsupervised Learning}

\author{Jayaraman~J.~Thiagarajan, Karthikeyan~Natesan~Ramamurthy and~Andreas~Spanias
\thanks{The authors are with the SenSIP Center, School of ECEE, Arizona State University, USA. 85287-5706}
\thanks{Email: \{jjayaram, knatesan, spanias\}@asu.edu}
}

\markboth{IEEE Transactions on Image Processing}%
{}
\maketitle


\begin{abstract}
In complex visual recognition tasks it is typical to adopt multiple descriptors, that describe different aspects of the images, for obtaining an improved recognition performance. Descriptors that have diverse forms can be fused into a unified feature space in a principled manner using kernel methods. Sparse models that generalize well to the test data can be learned in the unified kernel space, and appropriate constraints can be incorporated for application in supervised and unsupervised learning. In this paper, we propose to perform sparse coding and dictionary learning in the multiple kernel space, where the weights of the ensemble kernel are tuned based on graph-embedding principles such that class discrimination is maximized. In our proposed algorithm, dictionaries are inferred using multiple levels of $1-$D subspace clustering in the kernel space, and the sparse codes are obtained using a simple levelwise pursuit scheme. Empirical results for object recognition and image clustering show that our algorithm outperforms existing sparse coding based approaches, and compares favorably to other state-of-the-art methods.
\end{abstract}

\begin{IEEEkeywords}
Sparse coding, dictionary learning, multiple kernel learning, object recognition, clustering.
\IEEEpeerreviewmaketitle
\end{IEEEkeywords}

\section{Introduction}
\label{sec:intro}

\subsection{Combining Features for Visual Recognition}
Designing effective object recognition systems requires features that can describe salient aspects in an image, while being robust to variations within a class. Furthermore, adapting to variations in the visual appearance of images beyond the training set is crucial. As a result, recognition systems often employ feature extraction methods that provide high discrimination between the classes. However, it is observed that no feature descriptor can provide good discrimination for all classes of images. Hence, it is common to adaptively combine multiple feature descriptors that describe diverse aspects of the images such as color, shape, semantics, texture etc. The advantage of using multiple features in object recognition has been demonstrated in a number of research efforts \cite{sonnenburg2006large,rakotomamonjy2007more,gonen2011multiple,gehler2009feature,vedaldi2009multiple,jain2012spf, yang2012group}.  Another inherent challenge in visual recognition is the need to understand the intrinsic structure of high-dimensional features for improved generalization. Different assumptions on the data distribution will enable us to adapt suitable linear/non-linear models to analyze the features and thereby build more effective classifiers. Multiple kernel learning (MKL) is a well-known framework in computer vision that allows the use of multiple descriptors in a unified kernel space \cite{lanckriet2004learning}. The individual kernels used with each descriptor can be combined either using a non-negative linear combination or a Hadamard product or in any other manner, as long as the resulting kernel is positive semi-definite and thereby a valid kernel according to the Mercer theorem \cite{kernel}. An extensive review of the various approaches used to combine individual kernels can be found in \cite{gonen2011multiple}. Graph-embedding principles \cite{yan2007graph} can be integrated with multiple kernel learning to perform discriminative embedding in the unified space and the applications of this approach in supervised and unsupervised learning have been explored \cite{MKL}.

Since any symmetric positive definite kernel defines a unique reproducing kernel Hilbert space (RKHS) \cite{aronszajn1950theory}, any data vector in the space can be represented as a weighted combination of the training samples used to construct the kernel matrix. Since operations in the RKHS can be performed using just the kernel similarities without the knowledge of the actual form of the kernel, several pattern analysis and machine learning methods can be tractably solved using kernel methods. Since the kernel similarities can be measured using any suitable non-linear function, linear models learned in its RKHS can provide the power of non-linear models.

\begin{figure*}[t]
\begin{minipage}[b]{1.0\linewidth}
 \centering
\includegraphics[width = 15cm]{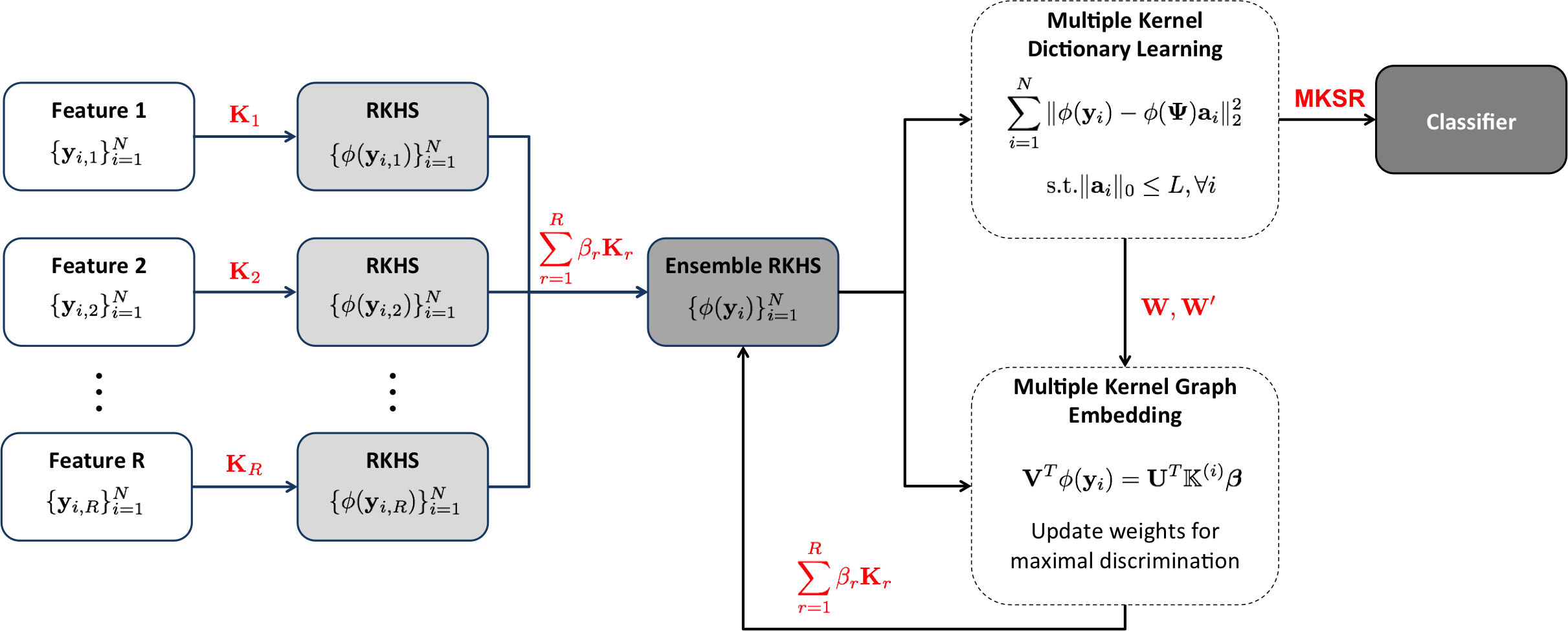}
\end{minipage}
\caption{The proposed multiple kernel sparse learning framework for multiclass object classification. Each image descriptor $\{\mathbf{y}_{i,r}\}_{i=1}^N$ defines an RKHS using the kernel $\mathbf{K}_r$, which are linearly combined using the non-negative weights, $\{\beta\}_{r=1}^R$. In the ensemble RKHS, multilevel dictionaries are optimized to provide sparse codes that can result in high discrimination between the classes.}
\label{Fig:arch}
\end{figure*}

\subsection{Sparse Coding in Classification and Clustering}
\label{sec:kernelmethods}
In the recent years, a variety of linear and non-linear modeling frameworks have been developed in the machine learning literature, aimed at exploiting the intrinsic structure of high-dimensional data. Sparse methods form an important class of models, where the data samples are approximated in a union of subspaces. Sparsity has been exploited in a variety of data processing and computer vision tasks such as compression \cite{Elad2006}, denoising \cite{elad2006image}, compressed sensing \cite{donoho2006compressed}, face classification \cite{wright}, blind source separation \cite{abolghasemi2012blind} and object recognition \cite{lcc}. The generative model for representing a data sample $\mathbf{y} \in \mathbb{R}^M$ using the sparse code $\mathbf{a} \in \mathbb{R}^K$ can be written as
\begin{equation}
\mathbf{y} = \mathbf{\Psi} \mathbf{a} + \mathbf{n},
\label{eqn:Sigrep}
\end{equation} where $\mathbf{\Psi}$ is the dictionary matrix of size $M \times K$ and $\mathbf{n}$ is the noise component not represented using the sparse code. Given the dictionary $\mathbf{\Psi}$, a variety of methods can be found in the literature to obtain sparse representations efficiently \cite{tropp}. When a sufficient amount of training data is available, the dictionary $\mathbf{\Psi}$ can be adapted to the data itself. The joint optimization problem of dictionary learning and sparse coding can be expressed as
\begin{equation}
\min_{\mathbf{\Psi},\mathbf{A}} \|\mathbf{Y} - \mathbf{\Psi} \mathbf{A}\|_F^2 + \lambda \sum_{i=1}^N\|\mathbf{a}_i\|_p \text{ subj. to } \forall k, \|\boldsymbol{\psi}_k\|_2 \leq 1,
\label{eqn:sparse_rep_dict_learn}
\end{equation}where the training data matrix $\mathbf{Y} = [\mathbf{y}_1,\ldots, \mathbf{y}_N]$, the coefficient matrix $\mathbf{A} = [\mathbf{a}_1,\ldots, \mathbf{a}_N]$, and $\|.\|_p$ is the $\ell_p$ norm ($0 \geq p \leq 1$) which measures the sparsity of the vector. Since (\ref{eqn:sparse_rep_dict_learn}) is not jointly convex, it is solved as an alternating optimization, where the dictionary and the sparse codes are optimized iteratively \cite{Elad2006,Lee2007}.  A wide range of dictionary learning algorithms have been proposed in the literature \cite{Rubin2010, mairal2012task}, some of which are tailored for specific tasks. The primary utility of sparse models with learned dictionaries in data processing stems from the fact that the dictionary atoms serve as \textit{predictive features}, capable of providing a good representation for some aspects of novel test data. From the viewpoint of statistical learning theory \cite{poggio2004}, a good predictive model is one that is generalizable, and dictionaries that satisfy this requirement have been proposed \cite{ramirez2012mdl, JT_MLD1}.

Sparse models have been used extensively in recognition and clustering frameworks. One of the first sparse coding based object recognition frameworks used codes obtained from raw image patches \cite{self_taught}. However, since then methods that use sparse codes of local descriptors aggregated to preserve partial spatial ordering have been proposed \cite{scspm, LC-KSVD, LLC} and they have achieved much better performance. In order to improve the performance further, algorithms that incorporate class-specific discriminatory information when learning the dictionary have been deveoped, and successfully applied for digit recognition and image classification \cite{mairal, bradley, LC-KSVD, zhang}. Improved discrimination can also be achieved by performing simultaneous sparse coding to enforce similar non-zero coefficient support \cite{Bengio,JT_subimage}, and ensuring that the sparse codes obey the constraints induced by the neighborhood graphs of the training data \cite{Laplacian, Ramamurthy2012}. By incorporating constraints that describe the underlying data geometry into dictionary learning, sparse models have been effectively used in non-linear manifold learning \cite{lpksvd} and activity recognition \cite{Rushil2013}.  Sparsity has also been shown to be useful in unsupervised clustering applications. In \cite{zheng2011graph}, the authors show that graph-regularized sparse codes cluster better when compared to using the data directly. Sparse coding graphs can be obtained by selecting the best representing subspace for each training sample, from the remaining samples \cite{Cheng2010} or from a dictionary \cite{ramirez2010classification} and subsequently used with spectral clustering.

Similar to other machine learning methods, adapting sparse representations to the RKHS has resulted in improved classifiers for computer vision tasks. Sparse models learned in the unified feature space often lead to discriminatory codes. Since the similarity function between the features in the RKHS is linear, samples that belong to the same class are typically grouped together in subspaces. Several approaches to perform kernel sparse coding have been proposed in \cite{kmp,kmp1,ksc}. In \cite{ksc}, the authors propose to learn dictionaries in the RKHS using a fixed point method. The well-known K-SVD \cite{Elad2006} and MOD learning algorithms have also been adapted to the RKHS, and an efficient object recognition system that combines multiple classifiers based on the kernel sparse codes is presented in \cite{Nguyen2013}. Sparse codes learned in the RKHS obtained by fusing intensity and location kernels have been successfully used for automated tumor segmentation \cite{JT_BIBE}.

\subsection{Proposed Multiple Kernel Sparse Learning}
\label{sec:mksl}
In this paper, we propose to perform dictionary learning and sparse coding in the multiple kernel space optimized for discriminating between various classes of images. Since the optimization of kernel weights is performed using graph-embedding principles, we also extend this approach to perform unsupervised clustering. Figure \ref{Fig:arch} illustrates our proposed approach for obtaining multiple kernel sparse representations (MKSR). As described earlier, generalization is a desired characteristic in learned dictionaries. Since the multilevel dictionary (MLD) learning \cite{JT_MLD1} has been shown to be generalizable to novel test data and stable with respect to perturbations of the training set, we choose to adapt this to the RKHS for creating kernel multilevel dictionaries. We learn the dictionaries by performing multiple levels of $1-$D subspace clustering in the RKHS obtained using the combination of multiple features. For novel test data, multiple kernel sparse codes can be obtained with a levelwise pursuit scheme that computes $1-$sparse representations in each level. In our setup, we construct the ensemble kernel as a non-negative linear combination of the individual kernels, and optimize the kernel weights using graph-embedding principles such that maximum discrimination between the classes is achieved. Since the graph affinity matrix for computing the embedding is based on the kernel sparse codes, this encourages the sparse codes to be discriminative across classes. The proposed algorithm iterates through the computation of embedding directions, updating of the weights, adaptation of the dictionary and calculation of new graph affinity matrices. Note that, the ensemble RKHS is modified in each iteration of the algorithm and a new dictionary is inferred for that kernel space. We empirically evaluate the use of the proposed framework in object recognition by using the proposed MKSR features with benchmark datasets (Oxford Flowers \cite{nilsback2006visual}, Caltech-101 \cite{caltech101}, and Caltech-256 \cite{caltech256}). Results show that our proposed method outperforms existing sparse-coding based approaches, and compares very well with other state-of-the-art methods. Furthermore, we extend the proposed algorithm to unsupervised learning and demonstrate performance improvements in image clustering.

\subsection{Paper Organization and Notation}
\label{sec:org}

In Section \ref{sec:ksr}, we provide a brief overview on sparse coding in the RKHS. Section \ref{sec:dict_learn_rkhs} presents a review of dictionary design principles, and summarizes MLD learning and kernel K-hyperline clustering. The kernel MLD algorithm and the pursuit scheme for obtain kernel sparse codes are also proposed in this Section. Section \ref{sec:mksr_dr} describes the proposed discriminative multiple kernel dictionary learning approach, which uses the KMLD and graph embedding to obtain discriminative dictionaries in the multiple kernel space. The effectiveness of the proposed algorithm is evaluated in object recognition (Section \ref{sec:obj_recog_eval}) and unsupervised clustering (Section \ref{sec:clustering_eval}) applications. 

A note on the notation used in this paper. We use uppercase bold letters for matrices, lowercase bold for vectors and non-bold letters for scalars and elements of matrices or vectors. The argument of the kernel $\phi(.)$, specifies whether it is a matrix or a vector.

\section{Sparse Representations in RKHS}
\label{sec:ksr}
Despite their great applicability, linear models in Euclidean spaces can be limited by their inability to exploit the non-linear relation between the image descriptors in visual recognition tasks. This limitation can be overcome using kernel functions that map the non-linearly separable descriptors into a high dimensional feature space, in which similar features are grouped together hence improving their linear separability \cite{kernel}. By choosing appropriate kernel functions that can extract task-specific information from the data, the learning problem can be effectively regularized. Since linear operations within the RKHS can be interpreted as non-linear operations in the input space, the linear models learned in the RKHS provide the power of a non-linear model. Furthermore, multiple kernel functions can be combined to create an ensemble RKHS thereby fusing the information from multiple descriptors.

Let us define the kernel function $\phi: \mathbb{R}^M \mapsto \mathcal{F}$, that maps the data samples from the input space to a
 RKHS $\mathcal{F}$. The data sample in the input space $\mathbf{y}$ transforms to $\phi(\mathbf{y})$ in the kernel space and therefore the $N$ training examples given by $\mathbf{Y} = [\mathbf{y}_1 \ldots \mathbf{y}_N]$ transform to $\phi(\mathbf{Y}) = [\phi(\mathbf{y}_1) \ldots \phi(\mathbf{y}_N)]$. The kernel similarity between the training examples $\mathbf{y}_i$ and $\mathbf{y}_j$ is defined using the pre-defined kernel function as  $K(\mathbf{y}_i,\mathbf{y}_j) \defeq \phi(\mathbf{y}_i)^T\phi(\mathbf{y}_j)$.  The dictionary in the RKHS is denoted by the matrix, $\phi(\mathbf{\Psi}) = [\phi(\boldsymbol{\psi}_1), \phi(\boldsymbol{\psi}_2),...,\phi(\boldsymbol{\psi}_K)]$, where each column indicates a dictionary element. The similarities between dictionary elements and the training examples can also be computed using the kernel function as  $\phi(\boldsymbol{\psi}_k)^T\phi(\mathbf{y}_j) = K(\boldsymbol{\psi}_k,\mathbf{y}_j)$ and $\phi(\boldsymbol{\psi}_k)^T\phi(\boldsymbol{\psi}_l) = K(\boldsymbol{\psi}_k,\boldsymbol{\psi}_l)$. Since all similarities can be computed exclusively using the kernel function, it is not necessary to know the transformation $\phi$. This greatly simplifies the computations in the feature space when the similarities are pre-computed, and this is known as the \textit{kernel trick}. We use the notation $\mathbf{K}_{\mathbf{Y}\mathbf{Y}} \in \mathbb{R}^{N \times N}$ to represent the matrix $\phi(\mathbf{Y})^T\phi(\mathbf{Y})$ and it contains the kernel similarities between all training examples. The similarity between two training examples, $K(\mathbf{y}_i,\mathbf{y}_j)$, is the $(i,j)^{\text{th}}$ element of $\mathbf{K}_{\mathbf{Y}\mathbf{Y}}$.

Sparse coding for a data sample $\mathbf{y}$ can be performed in the feature space as
\begin{equation}
\min_{\mathbf{a}} \|\phi(\mathbf{y}) - \phi(\mathbf{\Psi}) \mathbf{a} \|_2^2 + \lambda \|\mathbf{a}\|_0,
\label{eqn:ksc1}
\end{equation} where $\|.\|_0$ denotes the $\ell_0$ sparsity measure that counts the number of non-zero entries in a vector. The objective in (\ref{eqn:ksc1}) can be expanded as
\begin{align}
\nonumber &\phi(\mathbf{y})^T\phi(\mathbf{y}) - 2 \mathbf{a}^T\phi(\mathbf{\Psi})^T \phi(\mathbf{y}) + \mathbf{a}^T \phi(\mathbf{\Psi})^T \phi(\mathbf{\Psi})\mathbf{a} + \lambda \|\mathbf{a}\|_0, \\
& = K(\mathbf{y},\mathbf{y}) - 2 \mathbf{a}^T \mathbf{K}_{\mathbf{\Psi}\mathbf{y}} + \mathbf{a}^T \mathbf{K}_{\mathbf{\Psi} \mathbf{\Psi}} \mathbf{a} + \lambda \|\mathbf{a}\|_0.
\label{eqn:ksc2}
\end{align} Note that we have used the kernel trick here to simplify the computations. $\mathbf{K}_{\mathbf{\Psi}\mathbf{y}}$ is a $K \times 1$ vector containing the elements $K(\boldsymbol{\psi}_k,\mathbf{y})$, for $k = \{1, \ldots, K\}$ and $\mathbf{K}_{\mathbf{\Psi} \mathbf{\Psi}}$ is a $K \times K$ matrix containing the kernel similarities between all the dictionary atoms. Clearly, the objective function in (\ref{eqn:ksc2}) is similar to the sparse coding problem, except for the use of kernel similarities. As a result, existing algorithms can be easily extended to obtain kernel sparse codes. However, the computation of kernel matrices incurs additional complexity. Since any element in the RKHS lies in the span of the transformed training samples $\phi(\mathbf{Y})$, we can represent the dictionary $\phi(\mathbf{\Psi}) = \phi(\mathbf{Y}) \mathbf{C}$, where $\mathbf{C} \in \mathbb{R}^{N \times K}$. Hence, an alternate formulation for kernel sparse coding can be obtained.
\begin{equation}
\min_{\mathbf{a}} \|\phi(\mathbf{y}) - \phi(\mathbf{Y}) \mathbf{C} \mathbf{a} \|_2^2 + \lambda \|\mathbf{a}\|_0.
\label{eqn:ksc3}
\end{equation}

\section{Dictionary Design in RKHS}
\label{sec:dict_learn_rkhs}
Dictionaries designed in the Euclidean space have been useful in many image analysis and computer vision tasks. For example, the K-SVD algorithm \cite{Elad2006} trains a dictionary $\mathbf{\Psi}$ with $K$ columns such that the sparse codes $\mathbf{A}$ obtained minimize the reconstruction error for the $N$ training examples $\mathbf{Y}$. Clearly, other task specific constraints can also be posed in dictionary learning. For example, in \cite{Aviyente2006} the authors include the Fisher's discriminant penalty on the sparse codes, so that the sparse codes can discriminate well between the classes. A summary of the recent advances in dictionary learning can be found in \cite{tosic2011}.

In data processing and machine learning applications, the atoms in learned dictionaries serve as \textit{predictive features}, capable of representing different aspects of novel test data. As a result, some of the important questions to be considered when adapting sparse models to data include: (a) is the dictionary learning procedure stable? i.e., how sensitive is the learning algorithm to perturbations in the training set? and (b) does the learned dictionary generalize to novel test data? In \cite{ramirez2012mdl}, the authors proposed to improve the generalization of learned dictionaries by performing effective model selection. Furthermore, the multilevel dictionary (MLD) learning algorithm developed in \cite{JT_MLD1} provides an affirmative answer to the aforementioned questions using tools from statistical learning theory \cite{poggio2004}. In MLD learning, the representation of data $\mathbf{Y}$ is organized in multiple levels using a $1-$sparse representation for the sub-dictionary in each level. The sub-dictionaries are inferred using the K-hyperline (KHYPL) clustering algorithm \cite{Cichocki2009}, which is a $1-$D subspace clustering procedure. In summary, for each level a sub-dictionary is created and a residual matrix is obtained, which is used as the training data for the next level, and this process stops when a pre-defined number of levels or an error goal is reached.

Dictionary learning in the Euclidean space can be extended to any RKHS to identify predictive features in that space. The joint optimization of sparse coding and dictionary learning in the RKHS can be expressed as
\begin{equation}
\argmin_{\mathbf{A},\phi(\mathbf{\Psi})}\sum_{i=1}^N \|\phi({\mathbf{y}_i})-\phi(\mathbf{\Psi})\mathbf{a}_i\|_2^2+\lambda \sum_{i=1}^N \|\mathbf{a}_i\|_0.
\label{eqn:kernel_sc_dl}	
\end{equation} This can be solved as an alternating optimization, solving for sparse codes and the dictionary iteratively while fixing the other. As denoted in (\ref{eqn:ksc2}), the reconstruction error term can be expanded using the kernel trick, so that we only work with similarities in the RKHS and not the actual vectors themselves which may be very high dimensional. For some kernel functions, it is possible to learn the dictionary $\mathbf{\Psi}$ directly in the ambient space using the fixed point iteration method described in \cite{ksc}. Clearly, this is very restrictive since this method is limited to only some kernels and there are many useful kernels for which the even the kernel function cannot be expressed mathematically. The kernel K-SVD and kernel MOD methods proposed in \cite{Nguyen2013} express each dictionary element in the RKHS as the linear combination of the training data and uses an objective similar to (\ref{eqn:ksc3}) \cite{kernel}. The object recognition framework in \cite{Nguyen2013} use these dictionary learning schemes to optimize $\mathbf{C}$ separately for different classes of training data. For a test data sample, the sparse code is computed using kernel OMP (Orthogonal Matching Pursuit), and the sample is assigned to a class based on the dictionary which results in the minimum reconstruction error. Note that these procedures do not require that the mathematical form of any kernel function be known, rather they work directly on the kernel matrices.

In contrast, our method aims to infer a single dictionary for all classes, using an ensemble kernel and optimize the weights for linearly combining the kernels, such that maximal class discrimination is achieved. Since the MLD algorithm in the Euclidean space has been shown to be to good sparse predictive model \cite{JT_MLD1}, we propose to learn MLD in the kernel space using multiple levels of kernel K-hyperline (K2HYPL) clustering \cite{thiagarajan2012mixing}. We briefly describe the MLD algorithm below, and proceed to describe the main ingredients of our multiple kernel sparse representation (MKSR) method: the K2HYPL and the kernel MLD (KMLD) algorithms.

\subsection{Multilevel Dictionary Learning}
\label{sec:mld}
The multilevel dictionary is denoted as $\mathbf{\Psi} = [\mathbf{\Psi}_1 \mathbf{\Psi}_2  \ldots \mathbf{\Psi}_S]$, and the coefficient matrix is given as $\mathbf{A} = [\mathbf{A}_1^T \mathbf{A}_2^T \ldots \mathbf{A}_S^T]^T$, where $\mathbf{\Psi}_s$ is the sub-dictionary in level $s$ and $\mathbf{A}_s$ corresponds to the coefficient matrix in level $s$. The approximation in level $s$ of MLD learning is given as
\begin{equation}
\mathbf{R}_{s-1} = \mathbf{\Psi}_s\mathbf{A}_s+\mathbf{R}_{s}, \text{ for } s = {1,...,S},
\label{eqn:rep_layer_l}
\end{equation} where $\mathbf{R}_{s-1}$, $\mathbf{R}_{s}$ are the residuals for the levels $s-1$ and $s$ respectively and $\mathbf{R}_0 = \mathbf{Y}$. This implies that the residuals in level $s-1$ serve as the training data for level $s$. Note that the sparsity of the representation in each level is fixed at $1$. The optimization problem for each level of MLD learning is
\begin{align}
\nonumber &\argmin_{\mathbf{\Psi}_s,\mathbf{A}_s} \|\mathbf{R}_{s-1} - \mathbf{\Psi}_s\mathbf{A}_s\|_F^2 \text{ subject to } \|\mathbf{a}_{s,i}\|_0 \leq 1, \\
& \|\boldsymbol{\psi}_{s,k}\|_2 = 1 \text{ for } i = \{1, . . ., N\}, k = \{1, . . ., K\},
\label{eqn:opt_multilev_khyp}
\end{align} where $\mathbf{a}_{s,i}$ is the $i^{\text{th}}$ column in the coefficient matrix $\mathbf{A}_s$, and $\boldsymbol{\psi}_{s,k}$ is the $k^{\text{th}}$ column of the dictionary $\mathbf{\Psi}_{s}$.

The solution for (\ref{eqn:opt_multilev_khyp}) can be obtained using the KHYPL algorithm \cite{Cichocki2009}, a $1-$D subspace clustering procedure. This algorithm iteratively assigns the data samples to the closest $1-$D linear subspace, and updates the cluster centers to the best rank-$1$ approximation for all samples in a cluster. In order to simplify the notation, we will describe the procedure for the first level of MLD learning, and hence the training data will be $\mathbf{Y} = \mathbf{R}_0$. We will also denote the dictionary and coefficient matrices as $\mathbf{\Psi}$ and $\mathbf{A}$ respectively, dropping the subscript $s$. Note that the cluster centroids are the $1-$D subspaces represented by the dictionary atoms of $\mathbf{\Psi}$. In the cluster assignment stage, the $i^{\text{th}}$ training vector $\mathbf{y}_{i}$ is assigned to the $1-$D subspace $\boldsymbol{\psi}_{k}$ with which it has the least projection error. The cluster membership set $\mathcal{C}_k$ contains the indices of the training vectors assigned to the centroid $k$. In the centroid update stage, $\boldsymbol{\psi}_{k}$ is computed as the singular vector corresponding to the largest singular value of the set of training vectors assigned to the $k^{\text{th}}$ cluster centroid,  $\{\mathbf{y}_{i}|i \in \mathcal{C}_k\}$. Computing the principal singular vector can be linearized using a simple iterative scheme, which makes it straightforward to adapt this clustering method to the RKHS.

To express this clustering algorithm using matrix operations, we  define the membership matrix $\mathbf{Z} \in \mathbb{R}^{N \times K}$, where $z_{ik} = 1$ if and only if $i \in \mathcal{C}_k$. Cluster assignment is performed by computing
\begin{equation}
\mathbf{H} = \mathbf{Y}^T \mathbf{\Psi},
\label{eqn:A_comp}
\end{equation} and then setting $\mathbf{Z} = g(\mathbf{H})$, where $g(.)$ is a function that operates on a matrix and returns $1$ at the location of absolute maximum of each row and zero elsewhere. This is equivalent to locating the cluster centroid that has the minimum projection error or the maximum correlation for each training sample. Let us also define the coefficient matrix $\mathbf{A} = \mathbf{Z} \odot \mathbf{H}$, where $\odot$ indicates the Hadamard product. The centroid update can be then performed as
\begin{equation}
\mathbf{\Psi} = \mathbf{Y} \mathbf{A} \mathbf{\Gamma}(\mathbf{Y} \mathbf{A})^{-1}.
\label{eqn:centroid_comp}
\end{equation} Here, $\mathbf{\Gamma}(.)$ is a function that operates on a matrix and returns a diagonal matrix with the $\ell_2$ norm of each column in the matrix as its diagonal element. Therefore, $\mathbf{\Gamma}(\mathbf{Y} \mathbf{A})^{-1}$ ensures that the columns of $\mathbf{\Psi}$ are normalized.  Eqn. (\ref{eqn:centroid_comp}) obtains each cluster centroid as a normalized linear combination of the training samples associated with it. This is a linear approximation of the Power method for SVD computation which can be used to obtain the principal singular vector for each cluster. K-hyperline clustering is hence performed by iterating over membership update and centroid computation in (\ref{eqn:centroid_comp}).

\begin{table}[t]
\caption{The Kernel K-hyperline Clustering Algorithm.}
\centering
\begin{tabular}{|l|}
\hline
\textbf{Input} \\
$\mathbf{Y} = \left[ \mathbf{y}_i \right]_{i=1}^N$, $M \times N$ matrix of data samples. \\
$\mathbf{K}_{\mathbf{Y}\mathbf{Y}}$, $N \times N$ kernel matrix.\\
$K$, desired number of clusters.\\
\\
\textbf{Initialization}\\
- Randomly group data samples to initialize the membership matrix $\mathbf{Z}$.\\
- Based on $\mathbf{Z}$, obtain the rank-1 SVD for each cluster to initialize $\mathbf{\Psi}$.\\
- Compute the initial correlation matrix, $\mathbf{H} = \mathbf{Y}^T\mathbf{\Psi}$.\\ \\

\textbf{Algorithm}\\
\textbf{Loop until convergence}\\
\quad Loop for $L$ iterations\\
\quad \quad - Compute $\mathbf{A} = \mathbf{Z} \odot \mathbf{H}$. \\
\quad \quad - Compute $\mathbf{H} = \mathbf{K}_{\mathbf{Y}\mathbf{Y}} \mathbf{A} \mathbf{\Gamma}(\phi(\mathbf{Y}) \mathbf{A})^{-1}$.\\
\quad end \\
\quad - Update $\mathbf{Z}$ by identifying the index of absolute maximum in each  \\ \quad row of $\mathbf{H}$.\\
\textbf{end}\\
\hline
\end{tabular}
\label{Table:kkhypl}
\end{table}

\subsection{Kernel K-hyperline Clustering Algorithm}
\label{sec:k-khypl}
The KHYPL clustering can be performed in the RKHS with any valid kernel matrix, and we will refer to this as kernel K-hyperline (K2HYPL) clustering. Transformation of data to an appropriate feature space leads to tighter clusters and hence developing a kernel version of the K-lines clustering algorithm may lead to an improved clustering. Using the transformed data samples and dictionary elements, the correlation matrix, $\mathbf{H}$, in the RKHS can be computed in a manner similar to (\ref{eqn:A_comp}),
\begin{equation}
\mathbf{H} = \phi(\mathbf{Y})^T \phi(\mathbf{\Psi}),
\label{eqn:A_comp_feat}
\end{equation}and the membership matrix is given by $\mathbf{Z} = g(\mathbf{H})$. Hence, the cluster centers in the RKHS can be obtained as
\begin{equation}
\phi(\mathbf{\Psi}) =  \phi(\mathbf{Y}) \mathbf{A} \mathbf{\Gamma}(\phi(\mathbf{Y}) \mathbf{A})^{-1},
\label{eqn:centroid_comp_feat}
\end{equation}where $\mathbf{A} = \mathbf{Z} \odot \mathbf{H}$. The normalization term  is computed as
\begin{align}
\nonumber \mathbf{\Gamma}(\phi(\mathbf{Y}) \mathbf{A}) &=  diag((\phi(\mathbf{Y}) \mathbf{A})^T \phi(\mathbf{Y}) \mathbf{A})^{1/2} \\
&= diag(\mathbf{A}^T \mathbf{K}_{\mathbf{Y}\mathbf{Y}} \mathbf{A})^{1/2},
\label{eqn:norm_comp_feat}
\end{align} where $diag(.)$ is an operator that returns a diagonal matrix with the diagonal elements same as that of the argument matrix. Combining (\ref{eqn:A_comp_feat}), (\ref{eqn:centroid_comp_feat}) and (\ref{eqn:norm_comp_feat}), we obtain
\begin{align}
\nonumber \mathbf{H} &= \phi(\mathbf{Y})^T \phi(\mathbf{Y}) \mathbf{A} \mathbf{\Gamma}(\phi(\mathbf{Y}) \mathbf{A})^{-1}
 = \mathbf{K}_{\mathbf{Y}\mathbf{Y}} \mathbf{A} \mathbf{\Gamma}(\phi(\mathbf{Y}) \mathbf{A})^{-1}.
\end{align}The steps of this algorithm are presented in Table \ref{Table:kkhypl}. Note that initialization of the cluster centers is performed in the input space and not in the RKHS. The number of iterations in the inner loop, $L$, is fixed such that the coefficient estimate $\mathbf{H}$ converges.

\subsection{Proposed Kernel Multilevel Dictionary Learning Algorithm}
\label{sec:k-mld}
Given a set of training samples, our goal is to design multilevel dictionaries in the kernel space obtained using multiple kernels. The K2HYPL clustering procedure developed in the previous section can be used to learn the atoms in every level of the dictionary. In level $s$, we denote the sub-dictionary  by $\phi(\mathbf{\Psi}_s)$, the membership matrix by $\mathbf{Z}_s$, the coefficient matrix by $\mathbf{A}_s$, the input and the residual matrices by $\phi(\mathbf{Y}_s)$ and $\phi(\mathbf{R}_s)$ respectively. The training set for the first level is $\mathbf{Y}_1 = \mathbf{Y}$.

We begin by performing K2HYPL clustering in level $1$, which will yield the correlation matrix $\mathbf{H}_1 = \mathbf{K}_{\mathbf{Y}\mathbf{Y}} \mathbf{A}_1 \mathbf{D}_1$, where $\mathbf{D}_1 = \mathbf{\Gamma}(\phi(\mathbf{Y}_1) \mathbf{A}_1)^{-1} = diag(\mathbf{A}_1^T \mathbf{K}_{\mathbf{Y}\mathbf{Y}} \mathbf{A}_1)^{-1/2}$ indicates the diagonal matrix that normalizes the dictionary atoms of level $1$ in the kernel space. In KMLD learning, the residual vectors in a level are used as the training set to the next level. Hence, we compute the residuals as
\begin{align}
\nonumber \phi(\mathbf{R}_1) &= \phi(\mathbf{Y}_1) - \phi(\mathbf{\Psi}_1) \mathbf{A}_1^T = \phi(\mathbf{Y}_1) - \phi(\mathbf{Y}_1) \mathbf{A}_1 \mathbf{D}_1 \mathbf{A}_1^T,\\
&= \phi(\mathbf{Y}_1)\left[\mathbf{I}-\mathbf{A}_1 \mathbf{D}_1 \mathbf{A}_1^T\right] = \phi(\mathbf{Y}_2).
\label{eqn:residual1}
\end{align}Given the residuals from level $1$, the dictionary atoms in level $2$ can be computed as $\phi(\mathbf{\Psi}_2) = \phi(\mathbf{Y}_2) \mathbf{A}_2 \mathbf{D}_2$, where $\mathbf{D}_2 = diag\left((\phi(\mathbf{Y}_2) \mathbf{A}_2)^T(\phi(\mathbf{Y}_2) \mathbf{A}_2)\right)^{-1/2}$. This is simplified as
\begin{align}
\nonumber \mathbf{D}_2 = diag[\mathbf{A}_2^T &\left(\mathbf{I}-\mathbf{A}_1 \mathbf{D}_1 \mathbf{A}_1^T\right)^T \mathbf{K}_{\mathbf{Y}\mathbf{Y}} \\
& \left(\mathbf{I}-\mathbf{A}_1 \mathbf{D}_1 \mathbf{A}_1^T\right)\mathbf{A}_2]^{-1/2}.
\label{eqn:lev2_D}
\end{align}Similar to the previous level, the correlation matrix $\mathbf{H}_2$ is evaluated as
\begin{align}
\nonumber &\mathbf{H}_2 = \phi(\mathbf{Y}_2)^T \phi(\mathbf{\Psi}_2), \\
&= \left(\mathbf{I}-\mathbf{A}_1 \mathbf{D}_1 \mathbf{A}_1^T\right)^T \mathbf{K}_{\mathbf{Y}\mathbf{Y}}
\left(\mathbf{I}-\mathbf{A}_1 \mathbf{D}_1 \mathbf{A}_1^T\right)\mathbf{A}_2 \mathbf{D}_2.
\end{align} Table \ref{Table:kernelMLD} shows the detailed algorithm to learn a KMLD by generalizing the procedure for $S$ levels. Note that the innermost loop in the algorithm computes the cluster centroids using the linearized SVD procedure. The middle loop performs the K2HYPL clustering for a particular level.

\begin{table}[tb]
\caption{Kernel Multilevel Dictionary Learning algorithm.}
\small
\centering
\begin{tabular}{|l|}
\hline
\textbf{Input} \\
$\mathbf{K}_{\mathbf{Y}\mathbf{Y}}$, $N \times N$ kernel matrix for training data.\\
$K$, desired number of atoms per level. \\
$S$, total number of levels.\\
\\
\textbf{Algorithm}\\
\textbf{For $s = 1$ to $S$}\\
\quad- Randomly initialize the membership $\mathbf{Z}_s$ and compute the \\ 
\phantom{\quad- }initial correlation matrix, $\mathbf{H}_s$ for level $s$.\\
\quad\textbf{Loop until convergence}\\
\quad \quad - Loop for $L$ iterations\\
\quad \quad \quad - Compute $\mathbf{A}_s = \mathbf{Z}_s \odot \mathbf{H}_s$. \\
\quad \quad \quad - Compute $\displaystyle\mathbf{D}_s = diag\Bigg[\mathbf{A}_s^T \left(\prod_{t=1}^{s-1}(\mathbf{I}-\mathbf{A}_t \mathbf{D}_t \mathbf{A}_t^T)\right)^T$ \\
\quad \quad \quad \phantom{- Compute d} $\times \mathbf{K}_{\mathbf{Y}\mathbf{Y}} \left(\prod_{t=1}^{s-1}(\mathbf{I}-\mathbf{A}_t \mathbf{D}_t \mathbf{A}_t^T)\right) \mathbf{A}_s \Bigg]^{-1/2}$.\\
\quad \quad \quad - Evaluate $\displaystyle \mathbf{H}_s = \Bigg[\left(\prod_{t=1}^{s-1}(\mathbf{I}-\mathbf{A}_t \mathbf{D}_t \mathbf{A}_t^T)\right)^T \mathbf{K}_{\mathbf{Y}\mathbf{Y}}$ \\
\quad \quad \quad \phantom{- Evaluate test test}$\times \left(\prod_{t=1}^{s-1}(\mathbf{I}-\mathbf{A}_t \mathbf{D}_t \mathbf{A}_t^T)\right)\Bigg] \mathbf{A}_s \mathbf{D}_s$.\\
\quad \quad end \\
\quad \quad - Update $\mathbf{Z}_s$ using index of absolute maximum in each \\ 
\quad \quad \phantom{- }row of $\mathbf{H}_s$.\\
\quad\textbf{end}\\
\textbf{end}\\
- Overall multiple kernel sparse code matrix, $\mathbf{A} = [\mathbf{A}_1^T \ldots \mathbf{A}_S^T]^T$.\\
\hline
\end{tabular}
\label{Table:kernelMLD}
\end{table}

\subsection{Computing Sparse Codes for Test Data}
\label{sec:comp_sparse}
In this section, we describe a procedure to evaluate the sparse code for a novel test sample using the KMLD. The kernel matrix between the test sample $\mathbf{x}$ and the training data $\mathbf{Y}$ is given as $\mathbf{K}_{\mathbf{x}\mathbf{Y}}$. In order to obtain sparse codes for the test sample using the kernel MLD, we compute a sparse coefficient for each level using the dictionary atoms from that level. Similar to the training procedure, we first compute the correlations between the test sample and all dictionary elements in level $1$ as
\begin{equation}
\nonumber \boldsymbol{\alpha}_1 = \phi(\mathbf{x})^T \phi(\mathbf{\Psi}_1) = \phi(\mathbf{x})^T \phi(\mathbf{Y}_1) \mathbf{A}_1 \mathbf{D}_1 = \mathbf{K}_{\mathbf{x}\mathbf{Y}} \mathbf{A}_1 \mathbf{D}_1.
\label{eqn:al_test}
\end{equation}Following this, we determine the $1 \times K$ membership vector $\mathbf{z}_1 = g(\boldsymbol{\alpha}_1)$ and the coefficient vector $\mathbf{a}_1 = \mathbf{z}_1 \odot \boldsymbol{\alpha}_1$. The residual vector of the test sample can be computed as
\begin{align}
\nonumber \phi(\mathbf{r}_1) = \phi(\mathbf{x}) - \phi(\mathbf{\Psi}_1) \mathbf{a}_1^T = \phi(\mathbf{x}) - \phi(\mathbf{Y}_1) \mathbf{A}_1\mathbf{D}_1 \mathbf{a}_1^T.
\end{align}To determine a $1$-sparse code in level $2$, the residual vector $\mathbf{r}_1$ needs to be correlated with the dictionary atoms $\phi(\mathbf{\Psi}_2)$. Generalizing this to any level $s$, we can evaluate the correlations between the residual $\phi(\mathbf{r}_{s-1})$ and the dictionary atoms $\phi(\mathbf{\Psi}_s)$ as $\boldsymbol{\alpha}_s = \mathbf{M}_s \mathbf{A}_s \mathbf{D}_s$, where $\mathbf{M}_s$ is given by
\begin{align}
\nonumber & \left[\mathbf{K}_{\mathbf{x}\mathbf{Y}} - \sum_{t=1}^{s-1} \mathbf{a}_t \mathbf{D}_t \mathbf{A}_t^T \left(\prod_{p=1}^{t-1}(\mathbf{I}-\mathbf{A}_p \mathbf{D}_p \mathbf{A}_p^T)\right)^T \mathbf{K}_{\mathbf{Y}\mathbf{Y}}  \right] \\
&\phantom{\big[\mathbf{K}_{\mathbf{x}\mathbf{Y}} - \sum_{t=1}^{s-1} \mathbf{a}_t \mathbf{D}_t \mathbf{A}_t^T} \times \left[\left(\prod_{t=1}^{s-1}(\mathbf{I}-\mathbf{A}_t \mathbf{D}_t \mathbf{A}_t^T)\right)\right].
\label{eqn:al_M}
\end{align} The procedure for computing the sparse code for any test sample is given in Table \ref{Table:kernelMLD}.

\begin{table}[tb]
\caption{Sparse Codes for Test Data with KMLD.}
\small
\centering
\begin{tabular}{|l|}
\hline
\textbf{Input} \\
$\mathbf{K}_{\mathbf{x}\mathbf{Y}}$, $1 \times N$ kernel matrix for the test data $\mathbf{x}$.\\
$\{\mathbf{A}_s\}_{s=1}^S$, $K \times N$ coefficient matrices of training data. \\
$\{\mathbf{D}_s\}_{s=1}^S$, $K \times N$ normalization matrices of training data. \\
\\
\textbf{Algorithm}\\
\textbf{For $s = 1$ to $S$}\\
\quad - Compute $\boldsymbol{\alpha}_s = \mathbf{M}_s \mathbf{A}_s \mathbf{D}_s$ where $\mathbf{M}_s$ is given by (\ref{eqn:al_M}).\\
\quad - Sparse code for level $s$ is $\mathbf{a}_s = \mathbf{z}_s \odot \boldsymbol{\alpha}_s$, where $\mathbf{z}_s = g(\boldsymbol{\alpha}_s)$.\\
\textbf{end}\\
- Overall multiple kernel sparse code, $\mathbf{a} = [\mathbf{a}_1^T \mathbf{a}_2^T \ldots \mathbf{a}_S^T]^T$.\\
\hline
\end{tabular}
\label{Table:KMLD_testdata}
\end{table}

\section{Proposed Discriminative Multiple Kernel Dictionary Learning}
\label{sec:mksr_dr}
Given a set of training samples, our goal is to design multilevel dictionaries and obtain MKSR in the unified space obtained using multiple kernels, such that the data from the different classes are discriminated well. After initializing the graphs, and the kernel weights, the iterative optimization proceeds in four steps: (a) computing the discriminative low-dimensional subspaces for the unified kernel space using graph embedding, (b) optimizing for the kernel weights, (c) learning the dictionary using the modified ensemble kernel obtained with the updated weights, and (d) updating the inter-class and inter-class graphs using the multiple kernel sparse codes.

Given the $N$ training samples and $R$ kernels, the ensemble kernel matrix $\mathbf{K}$ is computed as,
\begin{equation}
\mathbf{K} = \sum_{r=1}^R \beta_r \mathbf{K}_{r}.
\label{eqn:ens_kernel}
\end{equation} where $\mathbf{K}_{r}$ is the kernel matrix corresponding to the $r^{\text{th}}$ descriptor. Note that we simplify the notation from the previous sections by dropping the data-dependent subscript for the kernel matrix. The vector $\boldsymbol{\beta} = [\beta_r]_{r=1}^R$ is initialized to have equal weights for all kernels. In order to learn the KMLD for obtaining kernel sparse codes, we need to construct an ensemble kernel by updating $\boldsymbol{\beta}$, that optimizes for maximal discrimination between the classes. To achieve this, we perform supervised graph embedding in the RKHS induced by the ensemble kernel and iteratively optimize $\boldsymbol{\beta}$ for discrimination (Section \ref{sec:disc_embed_weight_upd}). The inter-class and intra-class affinity matrices are initialized based on the local neighborhood of each sample in the first iteration. However, in the subsequent iterations, they are computed using the multiple kernel sparse codes. The initial affinity matrices denoted by $\mathbf{W} \in \mathbb{R}^{N \times N}$ and $\mathbf{W}' \in \mathbb{R}^{N \times N}$ are constructed by averaging the kernel-wise affinity matrices $\{\mathbf{W}^{(r)}\}_{r=1}^R$ and $\{\mathbf{W}'^{(r)}\}_{r=1}^R$. For the $r^{\text{th}}$ kernel, the elements of the kernel-wise affinity matrix are defined as
\begin{equation}
w_{ij}^{(r)} = 
\begin{cases}
1 & \text{if } \pi_i = \pi_j \text{ AND } j \in \mathcal{N}_{r,\tau}(i), \\
0 & \text{otherwise}.
\end{cases}
\label{eqn:intragr_init}
\end{equation}
\begin{equation}
w_{ij}'^{(r)} = 
\begin{cases}
1 & \text{if } \pi_i \neq \pi_j \text{ AND } j \in \mathcal{N}'_{r,\tau'}(i), \\
0& \text{otherwise}.
\end{cases}
\label{eqn:intergr_init}
\end{equation} Here $\pi_i$ is the class label for the $i^{\text{th}}$ sample, $\mathcal{N}_{r,\tau}(i)$ is the intra-class neighborhood of $\tau$ elements with respect to the $r^{\text{th}}$ kernel similarity measure. Similar definition applies to $\mathcal{N}'_{r,\tau'}(i)$ for the inter-class neighbors.

\subsection{Discriminative Embedding and Weight Update}
\label{sec:disc_embed_weight_upd}
Given the inter- and intra-class affinity matrices, we formulate a problem similar to the multiple kernel dimensionality reduction framework \cite{MKL} for computing the discriminative projection directions and updating the kernel weights.

Let us denote the $d$ embedding directions as the $d$ columns of the matrix $\mathbf{V}$. Since each column in $\mathbf{V}$ lies in the span of $\phi(\mathbf{Y})$, it can be expressed as $\mathbf{v} = \sum_{i=1}^N \phi(\mathbf{y}_i) u_i$. Therefore the low-dimensional projection can be expressed as 
\begin{equation}
\mathbf{v}^T \phi(\mathbf{y}_i) = \sum_{n=1}^N \sum_{r = 1}^R u_n \beta_r K_r(\mathbf{y}_n,\mathbf{y}_i) = \mathbf{u}^T \mathbb{K}^{(i)} \boldsymbol{\beta}
\label{eqn:mkl_proj}
\end{equation} where $\mathbf{u} = [u_1 \cdots u_N]^T \in \mathbb{R}^{N}$, $K_r(\mathbf{y}_n,\mathbf{y}_i)$ is the $(n,i)^{\text{th}}$ element of the kernel $\mathbf{K}_r$, and
\begin{align}
\mathbb{K}^{(i)} = \left[ \begin{array}{ccc}
K_1(\mathbf{y}_1,\mathbf{y}_i) & \cdots & K_R(\mathbf{y}_1,\mathbf{y}_i) \\
\vdots & \ddots & \vdots \\
K_1(\mathbf{y}_N,\mathbf{y}_i) & \cdots & K_R(\mathbf{y}_N,\mathbf{y}_i) \end{array} \right] \in \mathbb{R}^{N \times R}.
\end{align} Now using graph embedding principles \cite{yan2007graph}, the discriminative projection directions $\mathbf{V}$, or equivalently $\mathbf{U} = [\mathbf{u}_1 \ldots \mathbf{u}_d] \in \mathbb{R}^{d \times N}$, and the kernel weights can be jointly optimized as
\begin{gather}
\nonumber \min_{\mathbf{U},\boldsymbol{\beta}} \sum_{i,j=1}^N \|\mathbf{U}^T \mathbb{K}^{(i)} \boldsymbol{\beta}-\mathbf{U}^T \mathbb{K}^{(j)} \boldsymbol{\beta}\|_2^2 w_{ij}\\
\nonumber \text{subj. to } \sum_{i,j=1}^N \|\mathbf{U}^T \mathbb{K}^{(i)} \boldsymbol{\beta}-\mathbf{U}^T \mathbb{K}^{(j)} \boldsymbol{\beta}\|_2^2 w'_{ij} = 1,\\
\beta_r \geq 0 \text{ for } r = \{1,\ldots,R\}.
\label{eqn:comb_opt_U_beta}
\end{gather} Since direct optimization of (\ref{eqn:comb_opt_U_beta}) is hard, we perform an alternating optimization procedure to solve for $\mathbf{U}$ and $\boldsymbol{\beta}$ respectively. Here $w_{ij}$ and $w'_{ij}$ indicate the elements of the matrices $\mathbf{W}$ and $\mathbf{W}'$ respectively.

\textbf{Optimizing $\mathbf{U}$}: In order to optimize $\mathbf{U}$, we fix $\boldsymbol{\beta}$ and rewrite (\ref{eqn:comb_opt_U_beta}) as the trace ratio minimization 
\begin{gather}
\min_{\mathbf{U}} \frac{\text{trace}(\mathbf{U}^T \mathbf{S}_{\mathbf{W}}^{\boldsymbol{\beta}}\mathbf{U})} {\text{trace}(\mathbf{U}^T \mathbf{S}_{\mathbf{W}'}^{\boldsymbol{\beta}}\mathbf{U})},
\label{eqn:opt_U_TR}
\end{gather} where 
\begin{gather}
\mathbf{S}_{\mathbf{W}}^{\boldsymbol{\beta}} = \sum_{i,j=1}^N (\mathbb{K}^{(i)}-\mathbb{K}^{(j)}) \boldsymbol{\beta} \boldsymbol{\beta}^T (\mathbb{K}^{(i)}-\mathbb{K}^{(j)})^T w_{ij},\\
\mathbf{S}_{\mathbf{W}'}^{\boldsymbol{\beta}} = \sum_{i,j=1}^N (\mathbb{K}^{(i)}-\mathbb{K}^{(j)}) \boldsymbol{\beta} \boldsymbol{\beta}^T (\mathbb{K}^{(i)}-\mathbb{K}^{(j)})^T w_{ij}'.
\label{eqn:def_SWA}
\end{gather} The global optimal solution for (\ref{eqn:opt_U_TR}) can be obtained using the decomposed Newton's method provided in \cite{jia2009trace}.

\textbf{Optimizing $\boldsymbol{\beta}$}: The optimal value for $\boldsymbol{\beta}$ for a given $\mathbf{U}$ can be obtained by rewriting (\ref{eqn:comb_opt_U_beta}) as,
\begin{gather}
\nonumber \min_{\boldsymbol{\beta}} \boldsymbol{\beta}^T \mathbf{S}_{\mathbf{W}}^{\mathbf{U}} \boldsymbol{\beta}\\
\text{subj. to } \boldsymbol{\beta}^T \mathbf{S}_{\mathbf{W}'}^{\mathbf{U}} \boldsymbol{\beta} = 1, \boldsymbol{\beta} \geq \mathbf{0},
\label{eqn:opt_beta}
\end{gather} where
\begin{gather}
\mathbf{S}_{\mathbf{W}}^{\mathbf{U}} = \sum_{i,j=1}^N (\mathbb{K}^{(i)}-\mathbb{K}^{(j)}) \mathbf{U} \mathbf{U}^T (\mathbb{K}^{(i)}-\mathbb{K}^{(j)})^T w_{ij},\\
\mathbf{S}_{\mathbf{W}'}^{\mathbf{U}} = \sum_{i,j=1}^N (\mathbb{K}^{(i)}-\mathbb{K}^{(j)}) \mathbf{U} \mathbf{U}^T (\mathbb{K}^{(i)}-\mathbb{K}^{(j)})^T w_{ij}'.
\label{eqn:def_SWb}
\end{gather} The presence of $\boldsymbol{\beta} \geq \mathbf{0}$ in (\ref{eqn:opt_beta}) prevents solving it as a trace-ratio problem. However, the non-convex problem can be relaxed to a convex program using the auxiliary matrix $\mathbf{B} = \boldsymbol{\beta}\boldsymbol{\beta}^T \in \mathbb{R}^{R \times R}$ and introducing the relaxed constraint $\mathbf{B} \succeq \boldsymbol{\beta}\boldsymbol{\beta}^T$. The problem (\ref{eqn:opt_beta}) can then be posed as
\begin{gather}
\nonumber \min_{\boldsymbol{\beta}, \mathbf{B}} \text{trace} (\mathbf{S}_{\mathbf{W}}^{\mathbf{U}} \mathbf{B}) \text{ subj. to } \text{trace} (\mathbf{S}_{\mathbf{W}'}^{\mathbf{U}} \mathbf{B}) = 1,\\
\boldsymbol{\beta} \geq \mathbf{0}, 
\left[ \begin{array}{cc}
1 &  \boldsymbol{\beta}^T \\
\boldsymbol{\beta} & \mathbf{B} \end{array} \right] \succeq 0,
\label{eqn:opt_beta1}
\end{gather} and this can be solved efficiently using semi-definite programming. Using these optimization schemes $\mathbf{U}$ and $\boldsymbol{\beta}$ are iteratively optimized, and after this procedure has converged, the ensemble kernel $\mathbf{K}$ in (\ref{eqn:ens_kernel}) is recomputed using the updated $\boldsymbol{\beta}$. Given the updated ensemble kernel, we learn a new KMLD for multiple kernel sparse coding, using the procedure described in Table \ref{Table:kernelMLD}. Note that, in the subsequent iterations we will construct affinity matrices based on the MKSRs and use the procedure described above to update $\boldsymbol{\beta}$.

\begin{table}[tb]
\caption{Discriminative KMLD Learning Algorithm.}
\small
\centering
\begin{tabular}{|l|}
\hline
\textbf{Input} \\
$\{\mathbf{K}_r\}_{r=1}^R$, $R$ kernel matrices.\\
$K$, desired number of atoms per level. \\
$S$, total number of levels.\\
$d$, reduced dimension for the embedding.\\
$\tau$, $\tau'$, intra- and inter-class neighborhood size. \\
\\
\textbf{Initialization}\\
- Set the $R$ kernel weights $[\beta_r]_{r=1}^R$ to the same value. \\
- Initialize the graph affinities $\mathbf{W}$ and $\mathbf{W}'$ using (\ref{eqn:intragr_init}) and (\ref{eqn:intergr_init}).\\ \\
\textbf{Algorithm}\\
\textbf{Loop until convergence}\\
\quad \textbf{Loop until convergence}\\
\quad \quad- Update the discriminative embedding using (\ref{eqn:opt_U_TR}). \\
\quad \quad- Update the kernel weights using (\ref{eqn:opt_beta1}). \\
\quad \textbf{End}\\
\quad - Use (\ref{eqn:ens_kernel}) to compute ensemble kernel with updated $\boldsymbol{\beta}$.\\
\quad - Train a KMLD using the algorithm in Table \ref{Table:kernelMLD}.\\
\quad - Recompute graph affinities using (\ref{eqn:intragr_upd}) and (\ref{eqn:intergr_upd}).\\
\textbf{End}\\
\hline
\end{tabular}
\label{Table:disc_kmld}
\end{table}

\subsection{Updating the Intra- and Inter-Class Graphs}
\label{sec:inter_intra_class_graph}
Using the multiple kernel sparse codes $\mathbf{A}$ obtained from the KMLD algorithm, inter- and intra-class graphs are updated. The affinity matrices $\mathbf{W}$ and $\mathbf{W}'$ are given as
\begin{equation}
w_{ij} = 
\begin{cases}
|\mathbf{a}_i^T \mathbf{a}_j| & \text{if } \pi_i = \pi_j \text{ AND } j \in \mathcal{N}_{r,\tau}(i), \\
0 & \text{otherwise},
\end{cases}
\label{eqn:intragr_upd}
\end{equation}
\begin{equation}
w_{ij}' = 
\begin{cases}
|\mathbf{a}_i^T \mathbf{a}_j| & \text{if } \pi_i \neq \pi_j \text{ AND } j \in \mathcal{N}'_{r,\tau'}(i), \\
0 & \text{otherwise}.
\end{cases}
\label{eqn:intergr_upd}
\end{equation} In other words, we use the correlations between the resulting MKSRs to identify graph edges instead of the kernel-wise neighborhood. Since sparse models use a union of subspaces to represent data samples, this graph is non-local and hence less sensitive to perturbations in the data samples. The steps involved in the algorithm for obtaining discriminative multiple kernel dictionaries are detailed in Table \ref{Table:disc_kmld}.

\section{Object Recognition Evaluation}
\label{sec:obj_recog_eval}
In this section, we describe the set of image descriptors and the kernel functions considered for our simulations and present discussions on the recognition performance using the Oxford Flowers, Caltech-101, and Caltech-256 benchmark datasets. Given an appropriate distance function for each descriptor, we constructed kernel matrices as follows
\begin{equation}
\mathcal{K}(\mathbf{y}_i,\mathbf{y}_j) = \exp(-\gamma \rho(\mathbf{y}_i,\mathbf{y}_j))
\end{equation} where $\rho(.,.)$ is the distance function and the parameter $\gamma$ is fixed as the inverse of the mean of the pairwise distances.

\begin{table}[tb]
  \setlength{\tabcolsep}{10pt}
    \renewcommand*{\arraystretch}{1.1}
  \centering
  \caption{Comparison of the classification accuracies on the Oxford flowers dataset.}
    \begin{tabular}{|c|c|}
   \hline
   \textbf{Method} & \textbf{\% Accuracy} \\
    \hline
    \hline
	Nilsback \textit{et.al.} \cite{nilsback2006visual} & 81.3 \\    
    Bi \textit{et.al.} \cite{bi2004column}&84.8 \\
    Sonnenberg \textit{et.al.} \cite{sonnenburg2006large}&85.2 \\
	Gehler \textit{et.al.} \cite{gehler2009feature}&85.5 \\
	\hline
	\textbf{Proposed}&\textbf{86.3} \\
    \hline
    \end{tabular}%
  \label{Table:flowers}%
\end{table}

\noindent \textbf{Oxford Flowers Dataset:} This dataset consists of flower images belonging to $17$ different classes with $80$ images per class. For our experiment, we used $20$ images per class for training and the rest for testing. Following the procedure in \cite{gehler2009feature}, we used seven different descriptors and constructed kernels with them. Each kernel matrix is computed using a different descriptor, namely clustered HSV values, SIFT features on the foreground region, SIFT features on the foreground boundary and three matrices derived from color, shape and texture vocabularies. Details on this feature extraction process can be found in \cite{nilsback2006visual}. For learning the kernel MLD, we fixed the number of levels $S=8$ and the number of atoms in each level at $16$. Note that, the number of levels in KMLD learning corresponds to the desired sparsity in the representation of each sample in the unified feature space. In order to construct the affinity matrices $\mathbf{W}$ and $\mathbf{W}'$, we fixed the neighborhood size within the class to $8$ samples and between classes to $20$ samples. The number of embedding dimensions $d$ was fixed at $100$. We optimized $\boldsymbol{\beta}$ and the dictionary using the algorithm in Table \ref{Table:disc_kmld}. The resulting sparse codes were used to train a linear SVM. We repeated this experiment for $5$ different random splits of train and test sets, and report the average results in Table \ref{Table:flowers}. We observed that the proposed algorithm achieves an improved classification accuracy in comparison to the other approaches that combines features based on multiple kernel learning and boosting. 

\begin{table}[tb] 
  \setlength{\tabcolsep}{5pt}
  \renewcommand*{\arraystretch}{1.2}
  \centering
  \caption{Comparison of the classification accuracies on the Caltech-101 dataset.}
    \begin{tabular}{|c|c|c|c|c|c|c|}
   \hline
    \multirow{2}[0]{*}{\textbf{Method}} & \multicolumn{6}{|c|}{\textbf{\# Training samples per class}} \\
    \cline{2-7}
    & \textbf{5} & \textbf{10} & \textbf{15} & \textbf{20} & \textbf{25} & \textbf{30} \\
    \hline
    \hline
    Zhang \textit{et.al.} \cite{zhang2006}&46.6&55.8&59.1&62&-&66.2 \\
    Lazebnik \textit{et.al.} \cite{Lazebnik}&-&-&56.4&-&- &64.6\\
	Griffin \textit{et.al.} \cite{caltech256}&44.2&54.5&59&63.3&65.8&67.6\\
	Boiman \textit{et.al.} \cite{boiman}&-&-&61&-&- &69.1\\
	Gemert \textit{et.al.} \cite{Gemert}&-&-&-&-&- &64.16\\
	Yang \textit{et.al.} \cite{scspm}&-&-&67&-&- &73.2\\
	Wang \textit{et.al.} \cite{LLC}&51.15&59.77&65.43&67.74&70.16&73.44\\
	Aharon \textit{et.al.} \cite{Elad2006}&49.8&59.8&65.2&68.7&71&73.2\\
	Zhang \textit{et.al.} \cite{zhang}&49.6&59.5&65.1&68.6&71.1&73\\
	Jiang \textit{et.al.} \cite{LC-KSVD}&54&63.1&67.7&70.5&72.3&73.6\\
	Boureau \textit{et.al.} \cite{boureau2010learning}&-&-&-&-&-&77.3 \\
	Liu \textit{et.al.} \cite{liu2011defense}&-&-&-&-&-&74.2 \\
	Sohn \textit{et.al.} \cite{sohn2011efficient}&-&-&71.3&-&-&77.8 \\
	Goh \textit{et.al.} \cite{goh2012unsupervised}&-&-&71.1&-&-&78.9 \\
	Nguyen \textit{et.al.} \cite{Nguyen2013}&56.5&67.2&72.5&75.8&77.6&80.1 \\
	Duchenne \textit{et.al.} \cite{duchenne2011graph}&-&-&75.3&-&-&80.1 \\
	Gehler \textit{et.al.} \cite{gehler2009feature}&59.5&69.2&74.63&77.6&79.6&82.1 \\
	Feng \textit{et.al.} \cite{feng2011geometric}&-&-&70.3&-&-&82.6 \\
	Lin \textit{et.al.} \cite{MKL}&59.2&68.9&74.9&77.2&79.2&- \\
	Todovoric \textit{et.al.} \cite{todorovic2008learning}&-&-&72&-&-&83 \\
	Yang \textit{et.al.} \cite{yang2012group}&-&66.2&75.1&\textbf{81.5}&\textbf{83.7}&\textbf{84.6} \\
	\hline
    \textbf{Proposed}&\textbf{59.9}&\textbf{69.5}&\textbf{75.7}&79.7&80.8&82.9\\
    \hline
    \end{tabular}%
  \label{Table:Caltech101}%
\end{table}

\noindent \textbf{Caltech Datasets:} For our next set of experiments, we used the Caltech datasets which are important benchmark datasets for object recognition. The Caltech-101 dataset \cite{caltech101} consists of $9144$ images belonging to $101$ object categories and an additional class of background images. The number of images in each category varies roughly between $40$ and $800$. We resized all images to be no larger than $300 \times 300$ with the aspect ratio preserved. The Caltech-256 dataset \cite{caltech256} contains $30,607$ images in $256$ categories and its variability makes it extremely challenging in comparison to the Caltech-101 dataset. Following the common evaluation procedure for the Caltech-101 dataset, we trained the classifiers using $5, 10, 15, 20, 25$ and $30$ training images per class and evaluated the performance using upto $50$ images per class for testing. Performance is measured as the average classification accuracy per class, thus balancing the influence of classes with a large number of test samples.  Similarly for Caltech-256, following the standard procedure we evaluated the recognition performance with the number of training images fixed at $15, 30$, and $45$ images per class and tested for $25$ images per class. In both datasets, the reported performance was obtained by averaging over $10$ different random splits of train and test sets.

For both the Caltech datasets, we combined $39$ features used in \cite{gehler2009feature} and $9$ additional features from \cite{vedaldi2009multiple}. These features include the spatial pyramid, the PHOG shape, the region of covariance, the local binary pattern, the Vis+, geometric blur, PHOW gray/color, and self-similarity. We used appropriate distance functions to construct the kernel matrices. For example, $\chi^2$ distance was used for the PHOG descriptor, and the geodesic distance for the region of covariance descriptor. For the kernel MLD learning, the number of levels was fixed at $32$, and the number of atoms in each level was chosen to be $\{8, 8, 16, 16, 32, 32\}$ for each training case. In each iteration of our algorithm, we fixed the number of embedding dimensions $d = 150$ and optimized $\boldsymbol{\beta}$ for the best discrimination. For the case of Caltech-101, we compared the performance of our proposed method with state-of-the-art results obtained with several sparse coding based approaches and other methods that combine multiple features. Table \ref{Table:Caltech101} shows the results on the Caltech-101 dataset for the different training conditions. We observed that the proposed algorithm outperformed other sparse coding based approaches proposed in the literature, including the method in \cite{Nguyen2013} that combined multiple features for classification. Furthermore, by incorporating sparse coding into the dimensionality reduction framework proposed in \cite{MKL}, we improved the performance by $1.6\%$ at $N_{tr} = 25$. When compared to other non sparse-coding approaches, our method outperforms the existing methods when $N_{tr} < 20$ and compares well as $N_{tr}$ increases. For example, when $N_{tr} = 30$ our method achieves a classification accuracy of $82.9\%$ in comparison to $84.6\%$ accuracy obtained with group-sensitive multiple kernel learning \cite{yang2012group} and $83\%$ with the method proposed in \cite{todorovic2008learning}. For the Caltech-256 dataset, we repeated the experiment with the same set of features and parameters ($S = 32$ with $32$ atoms in each level). The results reported in Table \ref{Table:Caltech256} show that our proposed method outperforms the existing methods at $N_{tr} = 30$, and compares well with the state-of-the-art approaches in other cases.

\begin{table}[tb]
  \setlength{\tabcolsep}{10pt}
    \renewcommand*{\arraystretch}{1.1}
  \centering
  \caption{Comparison of the classification accuracies on the Caltech-256 dataset.}
    \begin{tabular}{|c|c|c|c|}
   \hline
    \multirow{2}[0]{*}{\textbf{Method}} & \multicolumn{3}{|c|}{\textbf{\# Training samples per class}} \\
    \cline{2-4}
    & \textbf{15} & \textbf{30} & \textbf{45}\\
    \hline
    \hline
   	Gemert \textit{et.al.} \cite{Gemert}&-&27.17&-\\
	Griffin \textit{et.al.} \cite{caltech256}&28.3&34.1&- \\
	Yang \textit{et.al.} \cite{scspm} &27.73&34.02&37.46 \\
	Guo \textit{et.al.} \cite{ksc} &29.77&35.67&38.61 \\
	Wang \textit{et.al.} \cite{LLC} &34.46&41.19&45.31 \\
	Feng \textit{et.al.} \cite{feng2011geometric}&35.18&43.17&47.32 \\
	Gehler \textit{et.al.} \cite{gehler2009feature}&34.2&45.8&- \\
	Todovoric \textit{et.al.} \cite{todorovic2008learning}&-&\textbf{49.5}&- \\
	\hline
	\textbf{Proposed}&\textbf{35.6}&46.9&\textbf{49.9}\\
    \hline
    \end{tabular}%
  \label{Table:Caltech256}%
\end{table}%

\section{Using the Proposed Framework for Unsupervised Learning}
\label{sec:clustering_eval}
An important feature of the proposed approach is that the graph embedding step in our algorithm can be replaced by several unsupervised, supervised and semi-supervised learning strategies. In this section, we report the performance of our method in clustering, that incorporates unsupervised graph embedding based on multiple kernel sparse codes. We employ an approach similar to kernel LPP (locality preserving projections) \cite{MKL} to learn an embedding and subsequently optimize $\boldsymbol{\beta}$ for obtaining MKSR. For the unsupervised graph embedding, we construct a single affinity matrix $\mathbf{W}$ and a degree matrix $\mathbf{\Delta}$ as follows. An undirected graph is defined, with the training samples as vertices, and the similarity between the neighboring training samples are coded in the affinity matrix $\mathbf{W} \in \mathbb{R}^{N \times N}$. We compute the affinity matrix $\mathbf{W} = |\mathbf{A}^T \mathbf{A}|$, where $\mathbf{A}$ is the matrix of multiple kernel sparse coefficients. Following this, we sparsify $\mathbf{W}$ by retaining only the $\tau$ largest similarities for each sample. We construct the degree matrix $\mathbf{\Delta}$ with each diagonal element containing the sum of the corresponding row or column of $\mathbf{W}$. The algorithm in Table \ref{Table:disc_kmld} can be readily extended to this case by replacing $\mathbf{W}$ and $\mathbf{W'}$ by $\mathbf{W}$ and $\mathbf{\Delta}$ respectively. Finally, the set of MKSR obtained with our algorithm are used to construct the graph for spectral clustering, by setting $\mathbf{W} = |\mathbf{A}^T \mathbf{A}|$ and retaining only $\tau$ largest entries for each sample. 

We evaluated the clustering performance of the kernel sparse coding-based graphs using a benchmark subset of the Caltech-101 dataset ($20$ classes) \cite{MKL}. Similar to the object recognition simulations, we constructed kernel matrices with different image descriptors. The set of features used in this experiment include SIFT-ScSPM \cite{scspm}, self similarity, PHOG, Gist, C2-SWP, C2-ML and geometric blur. Table \ref{Table:clus} compares the results of our method against using each feature separately with kernel LPP and K-Means clustering. Clearly, our method provides superior clustering performance in comparison to the MKL-LPP approach used in \cite{MKL} in terms of both clustering accuracy ($\%$ Acc) and normalized mutual information (NMI) measures.

\begin{table}[tb]
  \setlength{\tabcolsep}{5pt}
    \renewcommand*{\arraystretch}{1.1}
  \centering
  \caption{Comparison of the clustering performance on a subset of Caltech-101.}

    \begin{tabular}{|c|c|c|c|}
   \hline
    \textbf{Feature} & \textbf{Method}&\textbf{\% Acc} & \textbf{NMI}\\
    \hline
    \hline
    SIFT-ScSPM&\multirow{7}[0]{*}{Kernel-LPP \cite{MKL}}&58.6&0.63\\
    SS-ScSPM&&57.4&0.61\\
    PHOG&&47.2&50.4\\
    Gist&&42.8&0.48\\
    C2-SWP&&31.8&0.38\\
    C2-ML&&44.7&0.51\\
    GB&&54.7&0.61\\
    \hline
 All&MKL-LPP \cite{MKL}&72.6&0.74\\
 \textbf{All}&\textbf{Proposed}&\textbf{75.1}&\textbf{0.77}\\
   
    \hline
    \end{tabular}%
  \label{Table:clus}%
\end{table}

\section{Conclusions}
\label{sec:concl}
In this paper, we proposed an approach for learning dictionaries and obtaining sparse codes in a space defined by the linear combination of kernels from multiple descriptors. The dictionaries were obtained using multiple level of $1-$D subspace clustering, and the sparse codes were computed using a simple pursuit scheme. The weights of the individual kernels were tuned using graph embedding principles such that the discrimination across multiple classes is maximized. The proposed approach was used in object recognition and unsupervised clustering, and results showed that the performance compares favorably to other state-of-the-art methods. Since the framework is based on graph-embedding, it can be easily extended to other machine learning frameworks such as semi-supervised learning.

\bibliographystyle{IEEEtran}
\bibliography{refs}

\begin{thebibliography}{10}
\providecommand{\url}[1]{#1}
\csname url@samestyle\endcsname
\providecommand{\newblock}{\relax}
\providecommand{\bibinfo}[2]{#2}
\providecommand{\BIBentrySTDinterwordspacing}{\spaceskip=0pt\relax}
\providecommand{\BIBentryALTinterwordstretchfactor}{4}
\providecommand{\BIBentryALTinterwordspacing}{\spaceskip=\fontdimen2\font plus
\BIBentryALTinterwordstretchfactor\fontdimen3\font minus
  \fontdimen4\font\relax}
\providecommand{\BIBforeignlanguage}[2]{{%
\expandafter\ifx\csname l@#1\endcsname\relax
\typeout{** WARNING: IEEEtran.bst: No hyphenation pattern has been}%
\typeout{** loaded for the language `#1'. Using the pattern for}%
\typeout{** the default language instead.}%
\else
\language=\csname l@#1\endcsname
\fi
#2}}
\providecommand{\BIBdecl}{\relax}
\BIBdecl

\bibitem{sonnenburg2006large}
S.~Sonnenburg, G.~R{\"a}tsch, C.~Sch{\"a}fer, and B.~Sch{\"o}lkopf, ``Large
  scale multiple kernel learning,'' \emph{JMLR}, vol.~7, pp. 1531--1565, 2006.

\bibitem{rakotomamonjy2007more}
A.~Rakotomamonjy, F.~Bach, S.~Canu, and Y.~Grandvalet, ``More efficiency in
  multiple kernel learning,'' in \emph{Proc. of ICML}, 2007, pp. 775--782.

\bibitem{gonen2011multiple}
M.~Gonen and E.~Alpaydin, ``Multiple kernel learning algorithms,'' \emph{JMLR},
  vol.~12, pp. 2211--2268, 2011.

\bibitem{gehler2009feature}
P.~Gehler and S.~Nowozin, ``On feature combination for multiclass object
  classification,'' in \emph{Proc. of ICCV}, 2009, pp. 221--228.

\bibitem{vedaldi2009multiple}
A.~Vedaldi, V.~Gulshan, M.~Varma, and A.~Zisserman, ``Multiple kernels for
  object detection,'' in \emph{Proc. of ICCV}, 2009, pp. 606--613.

\bibitem{jain2012spf}
A.~Jain, S.~Vishwanathan, and M.~Varma, ``{SPF-GMKL}: generalized multiple
  kernel learning with a million kernels,'' in \emph{Proc. ACM SIGKDD}.\hskip
  1em plus 0.5em minus 0.4em\relax ACM, 2012, pp. 750--758.

\bibitem{yang2012group}
J.~Yang, Y.~Tian, L.-Y. Duan, T.~Huang, and W.~Gao, ``Group-sensitive multiple
  kernel learning for object recognition,'' \emph{IEEE TIP}, vol.~21, no.~5,
  pp. 2838--2852, 2012.

\bibitem{lanckriet2004learning}
G.~R. Lanckriet, N.~Cristianini, P.~Bartlett, L.~E. Ghaoui, and M.~I. Jordan,
  ``Learning the kernel matrix with semidefinite programming,'' \emph{JMLR},
  vol.~5, pp. 27--72, 2004.

\bibitem{kernel}
J.~Shawe-Taylor and N.~Cristianini, \emph{Kernel Methods for Pattern
  Analysis}.\hskip 1em plus 0.5em minus 0.4em\relax Cambridge, UK: Cambridge
  University Press, 2004.

\bibitem{yan2007graph}
S.~Yan, D.~Xu, B.~Zhang, H.-J. Zhang, Q.~Yang, and S.~Lin, ``Graph embedding
  and extensions: a general framework for dimensionality reduction,''
  \emph{IEEE TPAMI}, vol.~29, no.~1, pp. 40--51, 2007.

\bibitem{MKL}
Y.~Lin, T.~Liu, and C.~Fuh, ``Multiple kernel learning for dimensionality
  reduction,'' \emph{IEEE TPAMI}, vol.~33, no.~6, pp. 1147--1160, Jun. 2011.

\bibitem{aronszajn1950theory}
N.~Aronszajn, ``Theory of reproducing kernels,'' \emph{Trans. of the Amer.
  math. society}, vol.~68, no.~3, pp. 337--404, 1950.

\bibitem{Elad2006}
M.~Aharon, M.~Elad, and A.~Bruckstein, ``K-\textsc{SVD}: An algorithm for
  designing overcomplete dictionaries for sparse representation,'' \emph{IEEE
  TSP}, vol.~54, no.~11, pp. 4311--4322, 2006.

\bibitem{elad2006image}
M.~Elad and M.~Aharon, ``Image denoising via sparse and redundant
  representations over learned dictionaries,'' \emph{IEEE TIP}, vol.~15,
  no.~12, pp. 3736--3745, 2006.

\bibitem{donoho2006compressed}
D.~Donoho, ``Compressed sensing,'' \emph{IEEE Trans. Info. Theory}, vol.~52,
  no.~4, pp. 1289--1306, 2006.

\bibitem{wright}
J.~Wright, A.~Yang, A.~Ganesh, S.~S. Sastry, and Y.~Ma, ``Robust face
  recognition via sparse representation,'' \emph{IEEE TPAMI}, vol.~31, no.~2,
  pp. 210--227, 2009.

\bibitem{abolghasemi2012blind}
V.~Abolghasemi, S.~Ferdowsi, and S.~Sanei, ``Blind separation of image sources
  via adaptive dictionary learning,'' \emph{Image Processing, IEEE Transactions
  on}, vol.~21, no.~6, pp. 2921--2930, 2012.

\bibitem{lcc}
K.~Yu and T.~Zhang, ``Nonlinear learning using local coordinate coding,'' in
  \emph{Advances in NIPS}, 2009.

\bibitem{tropp}
J.~A. Tropp and S.~J. Wright, ``Computational methods for sparse solution of
  linear inverse problems,'' \emph{Proc. IEEE}, vol.~98, no.~6, pp. 948--958,
  2010.

\bibitem{Lee2007}
H.~Lee, A.~Battle, R.~Raina, and A.~Ng, ``Efficient sparse coding algorithms,''
  in \emph{Advances in NIPS}, 2007.

\bibitem{Rubin2010}
R.~Rubinstein, A.~Bruckstein, and M.~Elad, ``Dictionaries for sparse
  representation modeling,'' \emph{Proc. of the IEEE}, vol.~98, no.~6, pp.
  1045--1057, 2010.

\bibitem{mairal2012task}
J.~Mairal, F.~Bach, and J.~Ponce, ``Task-driven dictionary learning,''
  \emph{IEEE TPAMI}, vol.~34, no.~4, pp. 791--804, 2012.

\bibitem{poggio2004}
T.~Poggio, R.~Rifkin, S.~Mukherjee, and P.~Niyogi, ``General conditions for
  predictivity in learning theory,'' \emph{Nature}, vol. 428, no. 6981, pp.
  419--422, 2004.

\bibitem{ramirez2012mdl}
I.~Ram{\'\i}rez and G.~Sapiro, ``An {MDL} framework for sparse coding and
  dictionary learning,'' \emph{IEEE TSP}, vol.~60, no.~6, pp. 2913--2927, 2012.

\bibitem{JT_MLD1}
J.~J. Thiagarajan, K.~N. Ramamurthy, and A.~Spanias, ``Learning stable
  multilevel dictionaries for sparse representations,''
  \emph{http://arxiv.org/abs/1303.0448}, 2013.

\bibitem{self_taught}
R.~Raina, A.~Battle, H.~Lee, B.~Packer, and A.~Ng, ``Self-taught learning:
  Transfer learning from unlabeled data,'' in \emph{Proc. of ICML}, 2007.

\bibitem{scspm}
{J. Yang, K. Yu, Y. Gong and T. Huang}, ``Linear spatial pyramid matching using
  sparse coding for image classification,'' in \emph{Proc. of IEEE CVPR}, Jun.
  2009.

\bibitem{LC-KSVD}
Z.~Jiang, Z.~Lin, and L.~Davis, ``Learning a discriminative dictionary for
  sparse coding via label consistent {K-SVD},'' in \emph{{Proc. of IEEE CVPR}},
  2011.

\bibitem{LLC}
J.~Wang, J.~Yang, F.~Lv, T.~Huang, and Y.~Gong, ``Locality-constrained linear
  coding for image classification,'' in \emph{{Proc. of IEEE CVPR}}, 2010.

\bibitem{mairal}
J.~Mairal, F.~Bach, J.~Ponce, G.~Sapiro, and A.~Zisserman, ``Supervised
  dictionary learning,'' in \emph{Advances in NIPS}, 2009.

\bibitem{bradley}
D.~Bradley and J.~Bagnell, ``Differential sparse coding,'' in \emph{Advances in
  NIPS}, 2008.

\bibitem{zhang}
Q.~Zhang and B.~Li, ``Discriminative {K-SVD} for dictionary learning in face
  recognition,'' in \emph{{Proc. of IEEE CVPR}}, 2010.

\bibitem{Bengio}
S.~Bengio, F.~Pereira, Y.~Singer, and D.~Strelow, ``Group sparse coding,'' in
  \emph{Advances in NIPS}, 2009.

\bibitem{JT_subimage}
J.~J. Thiagarajan, K.~N. Ramamurthy, P.~Sattigeri, and A.~Spanias, ``Supervised
  local sparse coding of sub-image features for image retrieval,'' in
  \emph{Proc. of IEEE ICIP}, 2012.

\bibitem{Laplacian}
S.~Gao, I.~Tsang, L.~Chia, and P.~Zhao, ``Local features are not lonely –
  laplacian sparse coding for image classification,'' in \emph{Proc. of IEEE
  CVPR}, 2010.

\bibitem{Ramamurthy2012}
K.~N. Ramamurthy, J.~J. Thiagarajan, P.~Sattigeri, and A.~Spanias, ``Learning
  dictionaries with graph embedding constraints,'' in \emph{Proc. of Asilomar
  SSC}, 2012.

\bibitem{lpksvd}
Y.~Zhou, J.~Gao, and K.~Barner, ``Locality preserving k-svd for non-linear
  manifold learning,'' in \emph{Proc. of IEEE ICASSP}, 2013.

\bibitem{Rushil2013}
R.~Anirudh, K.~N. Ramamurthy, J.~J. Thiagarajan, P.~Turaga, and A.~Spanias, ``A
  heterogeneous dictionary model for representation and recognition of human
  actions,'' in \emph{Proc. of IEEE ICASSP}, 2013.

\bibitem{zheng2011graph}
M.~Zheng, J.~Bu, C.~Chen, C.~Wang, L.~Zhang, G.~Qiu, and D.~Cai, ``Graph
  regularized sparse coding for image representation,'' \emph{IEEE TIP},
  vol.~20, no.~5, pp. 1327--1336, 2011.

\bibitem{Cheng2010}
B.~Cheng, J.~Yang, S.~Yan, Y.~Fu, and T.~Huang, ``Learning with $\ell_1$-graph
  for image analysis,'' \emph{IEEE TIP}, vol.~19, no.~4, pp. 858--866, 2010.

\bibitem{ramirez2010classification}
I.~Ramirez, P.~Sprechmann, and G.~Sapiro, ``Classification and clustering via
  dictionary learning with structured incoherence and shared features,'' in
  \emph{Proc. of IEEE CVPR}, 2010, pp. 3501--3508.

\bibitem{kmp}
S.~Gou, Q.~Li, and X.~Zhang, ``A new dictionary learning method for kernel
  matching pursuit,'' in \emph{FSKD}, 2006, pp. 776--779.

\bibitem{kmp1}
P.~Vincent and Y.~Bengio, ``{Kernel Matching Pursuit},'' \emph{Machine
  Learning}, vol.~48, no. 1-3, pp. 165--187, 2002.

\bibitem{ksc}
S.~Gao, I.~Tsang, and L.~Chia, ``Kernel sparse representation for image
  classification and face recognition,'' in \emph{ECCV}, ser. Lecture Notes in
  Computer Science, 2010.

\bibitem{Nguyen2013}
H.~Nguyen, V.~Patel, N.~Nasrabad, and R.~Chellappa, ``Design of non-linear
  kernel dictionaries for object recognition,'' \emph{IEEE TIP}, vol.~PP,
  no.~99, 2013.

\bibitem{JT_BIBE}
J.~J. Thiagarajan, D.~Rajan, K.~N. Ramamurthy, D.~Frakes, and A.~Spanias,
  ``Automatic tumor identification using kernel sparse representations,'' in
  \emph{Proc. of IEEE BIBE}, 2012.

\bibitem{nilsback2006visual}
M.-E. Nilsback and A.~Zisserman, ``A visual vocabulary for flower
  classification,'' in \emph{Proc. of IEEE CVPR}, vol.~2, 2006, pp. 1447--1454.

\bibitem{caltech101}
L.~Fei-Fei, R.~Fergus, and P.~Perona, ``Learning generative visual models from
  few training examples: An incremental bayesian approach tested on 101 object
  categories,'' in \emph{Proc. of IEEE CVPR}, Jun. 2004.

\bibitem{caltech256}
\BIBentryALTinterwordspacing
G.~Griffin, A.~Holub, and P.~Perona, ``Caltech-256 object category dataset,''
  California Institute of Technology, Tech. Rep. 7694, 2007. [Online].
  Available: \url{http://authors.library.caltech.edu/7694}
\BIBentrySTDinterwordspacing

\bibitem{Aviyente2006}
K.~Huang and S.~Aviyente, ``Sparse representation for signal classification,''
  in \emph{Advances in NIPS}, 2006.

\bibitem{tosic2011}
I.~Tosic and P.~Frossard, ``Dictionary learning,'' \emph{IEEE Sig. Proc. Mag.},
  vol.~28, no.~2, pp. 27--38, 2011.

\bibitem{Cichocki2009}
Z.~He, A.~Cichocki, Y.~Li, S.~Xie, and S.~Sanei, ``{K}-hyperline clustering
  learning for sparse component analysis,'' \emph{Sig. Proc.}, vol.~89, pp.
  1011--1022, 2009.

\bibitem{thiagarajan2012mixing}
J.~J. Thiagarajan, K.~N. Ramamurthy, and A.~Spanias, ``Mixing matrix estimation
  using discriminative clustering for blind source separation,'' \emph{Dig.
  Sig. Proc.}, 2012.

\bibitem{jia2009trace}
Y.~Jia, F.~Nie, and C.~Zhang, ``Trace ratio problem revisited,'' \emph{IEEE
  TNN}, vol.~20, no.~4, pp. 729--735, 2009.

\bibitem{bi2004column}
J.~Bi, T.~Zhang, and K.~P. Bennett, ``Column-generation boosting methods for
  mixture of kernels,'' in \emph{Proc. of ACM SIGKDD}, 2004, pp. 521--526.

\bibitem{zhang2006}
H.~Zhang, A.~C. Berg, M.~Maire, and J.~Malik, ``Svm-knn: Discriminative nearest
  neighbor classification for visual category recognition,'' in \emph{Proc. of
  IEEE CVPR}, 2006, pp. 2126--2136.

\bibitem{Lazebnik}
{S. Lazebnik, C. Schmid and J. Ponce}, ``Beyond bags of features: Spatial
  pyramid matching for recognizing natural scene categories,'' in \emph{Proc.
  of IEEE CVPR}, Jun. 2006.

\bibitem{boiman}
O.~Boiman, E.~Shechtman, and M.~Irani, ``In defense of nearest-neighbor based
  image classification,'' in \emph{Proc. of IEEE CVPR}, Aug. 2008, pp. 1--8.

\bibitem{Gemert}
J.~C. Gemert, J.~Geusebroek, C.~Veenman, and A.~W. Smeulders, ``Kernel
  codebooks for scene categorization,'' in \emph{Proc. of ECCV}, 2008, pp.
  696--709.

\bibitem{boureau2010learning}
Y.-L. Boureau, F.~Bach, Y.~LeCun, and J.~Ponce, ``Learning mid-level features
  for recognition,'' in \emph{Proc. of IEEE CVPR}, 2010, pp. 2559--2566.

\bibitem{liu2011defense}
L.~Liu, L.~Wang, and X.~Liu, ``In defense of soft-assignment coding,'' in
  \emph{Proc. of IEEE CVPR}, 2011, pp. 2486--2493.

\bibitem{sohn2011efficient}
K.~Sohn, D.~Y. Jung, H.~Lee, and A.~O. Hero, ``Efficient learning of sparse,
  distributed, convolutional feature representations for object recognition,''
  in \emph{Proc. of ICCV}, 2011, pp. 2643--2650.

\bibitem{goh2012unsupervised}
H.~Goh, N.~Thome, M.~Cord, and J.-H. Lim, ``Unsupervised and supervised visual
  codes with restricted boltzmann machines,'' in \emph{Proc. of ECCV}, 2012,
  pp. 298--311.

\bibitem{duchenne2011graph}
O.~Duchenne, A.~Joulin, and J.~Ponce, ``A graph-matching kernel for object
  categorization,'' in \emph{Proc. of ICCV}, 2011, pp. 1792--1799.

\bibitem{feng2011geometric}
J.~Feng, B.~Ni, Q.~Tian, and S.~Yan, ``Geometric ℓ p-norm feature pooling for
  image classification,'' in \emph{Proc. of IEEE CVPR}, 2011, pp. 2609--2704.

\bibitem{todorovic2008learning}
S.~Todorovic and N.~Ahuja, ``Learning subcategory relevances for category
  recognition,'' in \emph{Proc. of IEEE CVPR}, 2008, pp. 1--8.

\end{thebibliography}
\end{document}